\documentclass[final,3p,times]{elsarticle}
\usepackage{graphicx}
\usepackage{amssymb}
\usepackage{amsmath}
\usepackage{mathtools}
\usepackage{bm}
\usepackage[hyphens]{url}
\usepackage{hyperref}
\usepackage{algorithm}
\usepackage{algpseudocode}
\usepackage{lineno}
\usepackage{booktabs}
\usepackage{adjustbox}
\usepackage{subcaption}
\usepackage{lipsum}
\usepackage{xcolor}

\DeclareMathOperator*{\argmin}{arg\,min}

\begin{document}

\begin{frontmatter}


\title{A parametric framework for kernel-based dynamic mode decomposition using deep learning}



\author[uvaaddress]{Konstantinos Kevopoulos}
\ead{konstantinos.kevopoulos@student.uva.nl}

\author[utaddress]{Dongwei Ye\corref{mycorrespondingauthor}}
\cortext[mycorrespondingauthor]{Corresponding authors}
\ead{d.ye-1@utwente.nl}

% \author[uvaaddress]{Mengwu Guo}
% \ead{m.guo@utwente.nl}

% \author[uvaaddress]{Alfons Hoekstra}
% \ead{a.g.hoekstra@uva.nl}

\address[uvaaddress]{Informatics Institute, Faculty of Science, University of Amsterdam, The Netherlands}
\address[utaddress]{Department of Applied Mathematics, University of Twente, The Netherlands}

% \author[uvaaddress,itmoaddress]{Pavel Zun}
% \ead{p.zun@uva.nl}

% \address[uvaaddress]{Computational Science Lab, Institute for Informatics, Faculty of Science, University of Amsterdam, The Netherlands}

% \address[itmoaddress]{National Center for Cognitive Research, ITMO University, Saint Petersburg, Russia}

\begin{abstract}
Surrogate modelling is widely applied in computational science and engineering to mitigate computational efficiency issues for the real-time simulations of complex and large-scale computational models or for many-query scenarios, such as uncertainty quantification and design optimisation. In this work, we propose a parametric framework for kernel-based dynamic mode decomposition method based on the linear and nonlinear disambiguation optimization (LANDO) algorithm. The proposed parametric framework consists of two stages, offline and online. The offline stage prepares the essential component for prediction, namely a series of LANDO models that emulate the dynamics of the system with particular parameters from a training dataset. The online stage leverages those LANDO models to generate new data at a desired time instant, and approximate the mapping between parameters and the state with the data using deep learning techniques. Moreover, dimensionality reduction technique is applied to high-dimensional dynamical systems to reduce the computational cost of training. Three numerical examples including Lotka-Volterra model, heat equation and reaction-diffusion equation are presented to demonstrate the efficiency and effectiveness of the proposed framework. 
\end{abstract}


\end{frontmatter}

% \linenumbers
% \numberwithin{table}{section}
% \numberwithin{figure}{section}

\section{Introduction}
Data-driven learning for dynamical systems is one of the important topics in scientific machine learning. Such learning processes characterise the latent behaviour of dynamical systems from available data and physics constraints, and formulate surrogates for predictive purposes. Data-driven learning methods can be mainly categorised into two types. The first type of methods spends effort in identifying the true underlying dynamics with \textit{a priori} knowledge. It is generally formulated as an inverse problem to infer physical parameters featuring the recognized governing equations \cite{ching2006bayesian,raissi2017machine,ye2024gaussian}. The estimation is not limited to the physical parameters but also the operators stemming from the spatial discretisation of a partial differential equation (PDE) \cite{peherstorfer2016data,geelen2023operator}. They are also closely relevant to input/parameter calibration in inverse uncertainty quantification from a stochastic perspective \cite{kennedy2001bayesian}. Alternatively, without prior knowledge of the governing equation, identification methods can be applied to recover the formulation by promoting the sparsity from a candidate library, e.g. sparse identification of nonlinear dynamics (SINDy) \cite{brunton2016discovering,kaiser2018sparse}. Such methods rely on the inclusiveness of the candidate library and are limited to linearity in parametrisation. The second type of methods approximates the dynamics from training data with a reparametrisation representation. For instance, deep learning-based techniques have been used as neural surrogates to represent the dynamics of the systems \cite{long2018pde,chen2018neural,kosmatopoulos1995high}.

Dynamic mode decomposition (DMD) is one of the state-of-the-art data-driven methods based on the linear tangent approximation of the dynamics over time \cite{schmid2010dynamic,proctor2016dynamic}. It was originally proposed in \cite{schmid2010dynamic} for mode stability analysis and model order reduction of fluid dynamics and has been extended in a wide range of other applications, such as neuroscience \cite{brunton2016extracting}, robotics \cite{berger2015estimation}, and plasma physics \cite{taylor2018dynamic}.  DMD considers that the state space of the dynamics can be approximated by the span of existing data as the basis functions, and subsequently expresses new predictions by the linear combination of those bases. \cite{tu2013dynamic} presented a different way to compute DMD modes, which can be viewed as a finite-dimensional data-driven approximation of the Koopman operator. However, it has been proved that the states themselves are not rich enough to approximate the Koopman operator in some scenarios \cite{tu2013dynamic}. Therefore the corresponding variants have been proposed, which are known as extended DMD \cite{williams2015data} and kernel DMD \cite{williams2014kernel} to achieve better estimation of Koopman eigenfunctions. Extended DMD enhances the bases by a chosen dictionary of states, while the kernel-based DMD further mitigates the dimensionality issue originated from the size of the explicit dictionary required in extended DMD with kernel functions. Baddoo et al. \cite{baddoo2022kernel} proposed the linear and nonlinear disambiguation optimization (LANDO) algorithm. It provides a unified perspective of the variants of DMD methods and enables a robust disambiguation of the underlying linear operator from nonlinear forcings in a system by the design of the kernel machine and sparse dictionary.

Data-driven models based on approximation methods are usually characterized by their specific model parameters, e.g. weights and bias in deep learning-based surrogate models or linear operators in DMD methods. Those model parameters are subjected to the tuning of data via the training process. In many-query scenarios, such as uncertainty quantification or design optimization, surrogate models are required for efficient and effective parametric predictions and are often constructed by considering those model parameters as functions of actual physical variables. A variety of methods have been proposed to extend DMD methods for such parametric predictions. Sayadi et al. \cite{sayadi2015parametrized} proposed to perform DMD on a snapshot matrix consisting of the snapshots with different parameter instances. However, the large stacked snapshot matrix causes computational efficiency issues and leads to the parameter-independent frequency, which affects the predictive performance in nonlinear problems. Huhn et al. \cite{huhn2023parametric} provided alternative algorithms by constructing individual DMD models for each parameter instance and interpolating eigen-pairs of those models over parameter space, thereby mitigating the aforementioned issues. \cite{sun2023parametric} leverages the radial basis function (RBF) network to interpolate snapshots over parameter space and construct DMD models based on the snapshots prediction from the RBF network. Similarly, \cite{andreuzzi2023dynamic} interpolated the predictions of DMD models trained with parameter instances to achieve the parametric prediction.

In this work, we focus on kernel-based DMD method based on the LANDO algorithm \cite{baddoo2022kernel} and extend the LANDO algorithm to parametric prediction using deep learning techniques. The LANDO algorithm aims to approximate the dynamics with kernel representation and optimises the weight matrix for kernel functions by minimising the residual between the prediction of dynamics and training data at each time step. To enable parametric prediction, individual LANDO models are trained with data for each known parameter instance and used for generating new data at a desired time instant. A deep neural network is subsequently applied to learn the mapping between the parameters and the states for that time instant. The proposed parametric framework is motivated by the parametric algorithm proposed in \cite{andreuzzi2023dynamic}. We distinguish the scenarios for low-dimensional and high-dimensional systems and employ a model order reduction technique, i.e. proper orthogonal decomposition, for high-dimensional systems to further improve the computational efficiency. Three numerical examples, including Lotka-Volterra model, heat equation and reaction-diffusion equation are demonstrated to showcase the computational effectiveness and efficiency of the proposed methods. The remaining part of the work is organized as follows. The problem statement and LANDO algorithm are detailed in Section~\ref{sec:ps} and \ref{sec:LANDO}. The parametric extension is presented in Section~\ref{sec:pLANDO}. Three numerical examples are demonstrated in Section~\ref{sec:example}. The discussion and summary of the work are presented Section~\ref{sec:discussion}.

\section{Problem statement}\label{sec:ps}
Consider a parametric dynamical system,
\begin{equation}\label{eq:ds}
    \frac{\mathrm{d}\mathbf{x}(t,\bm{\mu})}{\mathrm{d}t} = \mathbf{F}(\mathbf{x}(t,\bm{\mu}),\bm{\mu}),
\end{equation}
where $\mathbf{x}(t,\bm{\mu})\in \mathbb{R}^N$ denotes the state of the system evolving over time $t$ based on the dynamics $\mathbf{F}$. The dynamics of the system is characterised by the physical parameter $\bm{\mu}$ from the interested parameter space $\mathbb{P}\subset \mathbb{R}^{N_p}$. Alternatively, the time discrete form can be written as,
\begin{equation}\label{eq:ds_discrete}
    \mathbf{x}_{j+1}(\bm{\mu}) = \mathbf{F}(\mathbf{x}_j(\bm{\mu}),\bm{\mu}),
\end{equation}
where $\mathbf{F}: \mathbb{R}^N \rightarrow \mathbb{R}^N$ represents an iteration map that characterises the time evolution of the state from $\mathbf{x}_j$ to $\mathbf{x}_{j+1}$. We are interested in constructing a parametric surrogate model that approximates the mapping $(t,\bm{\mu}) \mapsto \bm{x}(t,\bm{\mu})$ based on training data $\{\bm{\mu}_i,\mathbf{X}_i\}_{i=1}^{N_{\mu}}$, where $N$ denotes the number of training data and $\mathbf{X}_i = [\mathbf{x}(t_1,\bm{\mu}_i) \,|\, \mathbf{x}(t_2,\bm{\mu}_i) \,| \cdots |\, \mathbf{x}(t_{N_t},\bm{\mu}_i)] \in \mathbb{R}^{N\times N_{t}}$ consisting of the snapshots collected at $N_t$ time instants. Prediction of the states $\bm{x}(t^*,\bm{\mu}^*)$ can be subsequently performed with given parameter $\bm{\mu}^*$ at a given time instant $t^*$. Note that, the snapshots over time in matrices $\mathbf{X}_i$ are not necessary to be collected/measured at the same time instants $\{t_i\}_{i=1}^{N_t}$ (or for the same number of time instants $N_t$) over different parameters $\{\bm{\mu}_i\}_{i=1}^{N_{\mu}}$. Without loss of generality, the snapshots in this work are presumed to be collected at same time instants $\{t_i\}_{i=1}^{N_t}$ for notation simplification, and it is straightforward to extend them to the more general scenario.

\section{LANDO algorithm}\label{sec:LANDO}
\subsection{Kernel learning for dynamical systems}
The LANDO algorithm was proposed in \cite{baddoo2022kernel} and aims to approximate the dynamics $\mathbf{F}$ with kernel method. Consider equation~\eqref{eq:ds} with fixed parameters $\bm{\mu}$ and the problem, therefore, deteriorates to learning a nonparametric dynamical system where the dynamics $\mathbf{F}$ only depends on the states $\mathbf{x}$. A surrogate model $\mathbf{f}(\mathbf{x})$ for the dynamics of system $\mathbf{F}$ can be constructed with an appropriate kernel function $k$,
\begin{equation}\label{eq:kernel}
   \mathbf{F}\approx \mathbf{f}(\mathbf{x}) = \sum_{i=1}^{N_t} \mathbf{w}_i k(\mathbf{x}_i, \mathbf{x}) = \mathbf{W} k(\mathbf{X}, \mathbf{x}),
\end{equation}
where $\mathbf{X} = [ \mathbf{x}_1 \, | \cdots | \,\mathbf{x}_{N_t} ] \in \mathbb{R}^{N\times N_t}$ denotes the snapshots matrix available for training and matrix $\mathbf{W} = [\mathbf{w}_1 \,| \cdots | \, \mathbf{w}_{N_t} ] \in \mathbb{R}^{N\times N_t}$ collects the weight vectors $\mathbf{w}_i$ corresponding to each kernel function. The choice of the kernel can be used to reflect prior knowledge of the system, such as specific symmetries, conservation laws or geometrical features. Those additional physics constraints can be integrated into the design of a kernel, effectively reducing the data requirements for training and improving the prediction performance \cite{baddoo2023physics}. Such kernel representation also provides a unified perspective of different DMD methods. For example, the exact DMD can be recovered by choosing a special case of linear kernel \cite{tu2013dynamic}.

In continuous time formulation \eqref{eq:ds}, the corresponding data $\mathbf{Y}$ for dynamics $\mathbf{F}$ are typically generated from numerical differentiation such as finite different methods, while in the discrete form, the data are taken from the snapshots of subsequent time steps $\mathbf{Y} =  [ \mathbf{x}_2 \, | \cdots | \,\mathbf{x}_{N_t+1} ]$. Therefore, the weight matrix can be computed via the optimization problem to minimise the residual between LANDO prediction and the actual data,
\begin{equation}\label{eq:optimziation_lando}
\underset{\mathbf{W}}{\operatorname{argmin}} \|\mathbf{Y}-\mathbf{W}k(\mathbf{X}, \mathbf{X})\|_{F}+\lambda R(\mathbf{W}),
\end{equation}
where $\|\cdot\|_{F}$ denotes the Frobenius norm and $\lambda R(\mathbf{W})$ is a proper regularisation. In the absence of the regulariser $\lambda R(\mathbf{W})$, the optimisation problem can be solved by the Moore-Penrose pseudoinverse of the kernel matrix as,
\begin{equation}\label{eq:pseudoinv_sol}
    \mathbf{W} = \mathbf{Y}k(\mathbf{X}, \mathbf{X})^{\dagger}.
\end{equation}
However, the computational cost of such an inversion operation can potentially be high due to the $N_t$ number of snapshots required to sufficiently learn nonlinear system dynamics. Additionally, the kernel matrix $k(\mathbf{X}, \mathbf{X})$ often has a high condition number and suffers from 
overfitting. To mitigate the aforementioned issues, sparse dictionary learning \cite{engel2004kernel} was applied to extract the most informative snapshots with the almost linearly dependent (ALD) test.

\subsection{Sparse dictionary via ALD test}\label{sec:sdl}
The ALD test constructs a sparse dictionary of snapshots recursively by measuring the distance between a query snapshot and the linear span of the current dictionary in feature space. Consider the $\mathbf{x}_1$ as the first snapshot of the sparse dictionary $\tilde{\mathbf{X}}$. In each iteration, a new candidate snapshot $\mathbf{x}_c$ from $\mathbf{X}$ is chosen and tested on how well it can be approximated in the feature space by the current dictionary,
\begin{equation}\label{sparsity_test}
\delta=\min _{\bm{\pi}}\left\|\boldsymbol{\phi}(\mathbf{x}_c)-\tilde{\boldsymbol{\Phi}} \boldsymbol{\pi}\right\|_2^2 ,
\end{equation}
where $\boldsymbol{\phi}(\mathbf{x}_{c})$ and $\tilde{\boldsymbol{\Phi}} = \bm{\phi}(\tilde{\mathbf{X}})$ refer to the candidate snapshot and the current sparse dictionary in feature space, respectively. The quantity $\delta$ refers to the minimum difference between the candidate snapshots and the linear span of the current dictionary with the minimiser $\boldsymbol{\pi}$. When $\delta$ is below a user-defined threshold, it indicates that the candidate can be well approximated by the current dictionary, otherwise the current feature space is not sufficiently rich to approximate the candidate and therefore this candidate is required to be added to the dictionary. Kernel methods operate with implicit feature space by kernel functions, and therefore the expression \eqref{sparsity_test} can be also rewritten as, 
\begin{equation}\label{delta}
    \delta = k(\mathbf{x}_c, \mathbf{x}_c) - \tilde{\mathbf{k}}^* \boldsymbol{\pi}, 
\end{equation}
where $\tilde{\mathbf{k}}^*$ is the conjugate transpose of $\tilde{\mathbf{k}} = k(\tilde{\mathbf{X}}, \mathbf{x}_c)$ and $\boldsymbol{\pi} = \tilde{\mathbf{K}}^{-1} \tilde{\mathbf{k}}$ is the minimiser, where $\tilde{\mathbf{K}} = k(\tilde{\mathbf{X}}, \tilde{\mathbf{X}})$. If the candidate sample is accepted, the kernel matrices will be updated in each iteration until the stopping criterion is fulfilled. Algorithm~\ref{alg:lando} demonstrates the entire process of the construction of the sparse dictionary. Additionally, to resolve the possible large condition number of the kernel matrices, Cholesky decomposition is applied for updating \cite{baddoo2022kernel}. We invite readers to \cite{baddoo2022kernel} for further details on the LANDO algorithm.

After the sparse dictionary has been formed, the coefficient matrix $\tilde{\mathbf{W}}$ of \eqref{eq:optimziation_lando} can be subsequently computed by 
\begin{equation}\label{eq:pseudoinv_sol_lando}
    \tilde{\mathbf{W}} = \mathbf{Y}k(\mathbf{\tilde{X}}, \mathbf{X})^{\dagger},
\end{equation}
As a result, two quantities define the surrogate model $\mathbf{f}(\mathbf{x})$, the sparse dictionary of samples $\mathbf{\tilde{X}}$, and the corresponding coefficient matrix $\tilde{\mathbf{W}}$.
Recall that expression \eqref{eq:pseudoinv_sol} involves $\mathbf{X} \in \mathbb{R}^{N \times N_t} \text{ and } \mathbf{W} \in \mathbb{R}^{N \times N_t}$ and after the sparse dictionary learning, the dimension of the matrices are reduced to $\tilde{\mathbf{X}} \in \mathbb{R}^{N \times m} \text{ and } \tilde{\mathbf{W}} \in \mathbb{R}^{N \times m}$, where $m$ denotes the total number of snapshots in the sparse dictionary and $m \ll N_t$.

\begin{algorithm}[tb!]
\caption{Sparse dictionary learning}
\label{alg:lando}
\begin{algorithmic}
\Require Snapshot matrix $\mathbf{X}$, kernel function $k$, sparsity threshold $\nu$
\Ensure  Sparse dictionary $\tilde{\mathbf{X}}$
\vspace{0.1cm}

\State $\mathbf{X} \gets \text{Perm}(\mathbf{X})$ \Comment{\textcolor{gray}{Randomly permute the columns of $\mathbf{X}$}}
\State $\tilde{\mathbf{X}} \gets \mathbf{x}_1$ \Comment{\textcolor{gray}{Take $\mathbf{x}_1$ as the initial $\tilde{\mathbf{X}}$}}
\For{$i=2,\cdots,N_t$} 
    \State $\mathbf{x}_{c} \gets \mathbf{x}_i$
    \State  $\tilde{\mathbf{k}} \gets k(\tilde{\mathbf{X}}, \mathbf{x}_c)$
    \State  $\tilde{\mathbf{K}} \gets k(\tilde{\mathbf{X}}, \tilde{\mathbf{X}})$
    \State  $\boldsymbol{\pi} \gets \tilde{\mathbf{K}}^{-1} \tilde{\mathbf{k}}$
    \State $\delta \gets k(\mathbf{x}_c, \mathbf{x}_c) - \tilde{\mathbf{k}}^* \boldsymbol{\pi}$

    \If {$\delta \leq \nu$}
        \State $\tilde{\mathbf{X}} \gets \tilde{\mathbf{X}}$ \Comment{\textcolor{gray}{The dictionary is not updated}}
    \Else
        \State $\tilde{\mathbf{X}} \gets [\tilde{\mathbf{X}} \,| \, \mathbf{x}_c ]$ 
    \Comment{\textcolor{gray}{The dictionary is updated}}
    \EndIf
\EndFor

\end{algorithmic}
\end{algorithm}

\section{Parametric framework of LANDO}\label{sec:pLANDO}
\begin{figure}
    \centering
    \includegraphics[width=\linewidth]{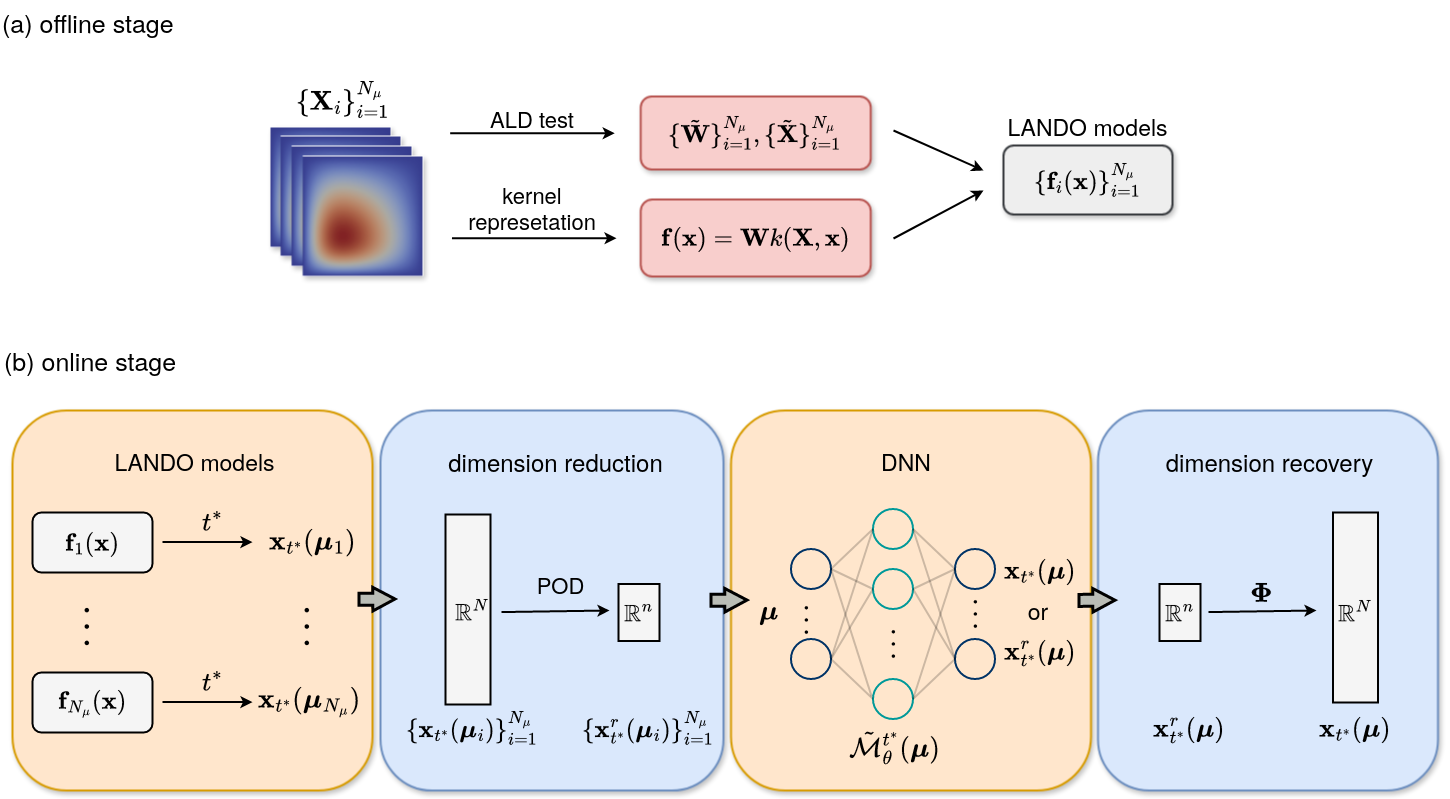}
    \caption{A schematic diagram for proposed parametric LANDO framework. At the offline stage, a series of LANDO models are prepared and used for data generation at particular time instant $t^*$ during online stage. A DNN surrogate model $\tilde{\mathcal{M}}^{t^*}_{\bm{\theta}}(\bm{\mu})$ is subsequently applied to learn the mapping $\mathcal{M}^{t^*}(\bm{\mu})$ between parameters and states based on those data. For high-dimensional dynamical systems, dimensionality reduction technique based on POD is employed to the state data before DNN learning and recover to the full state after DNN prediction (demonstrated in blue boxes).}
    \label{fig:diagram}
\end{figure}
The parametric LANDO proposed in this work aims to emulate efficiently the parametric dynamical system described by \eqref{eq:ds} or \eqref{eq:ds_discrete}. Without loss of generality, we present the parametric framework with the continuous form of the system as shown with \eqref{eq:ds}. In particular the data $\mathbf{Y}_i$ used for optimization in \eqref{eq:optimziation_lando} are computed from the time derivative of states using numerical differentiation, e.g. finite difference method, based on $\mathbf{X}_i$ corresponding to the parameter instance $\bm{\mu}_i$. A schematic diagram of the parametric LANDO framework is presented in Figure~\ref{fig:diagram}.

\subsection{Offline stage}
Given the training data $\{\bm{\mu}_i,\mathbf{X}_i\}_{i=1}^{N_{\mu}}$, the parametric framework starts from constructing an individual LANDO surrogate model for each parameter instance $\bm{\mu}_i\in \mathbb{P}$, for $i=1,\cdots,N_{\mu} $, which is denoted as,

\begin{equation}\label{eq:p_kernel}
   \mathbf{f}_i(\mathbf{x})  = \mathbf{f}(\mathbf{x},\bm{\mu}_i) = \sum_{j=1}^{m} \tilde{\mathbf{w}}_{i,j} k(\tilde{\mathbf{x}}_{i.j}, \mathbf{x}) = \tilde{\mathbf{W}}_i  \, k(\tilde{\mathbf{X}}_i, \mathbf{x}),
\end{equation}
where $\tilde{\mathbf{x}}_{i,j}$ denotes the $j$th column of the sparse dictionary $\tilde{\mathbf{X}}_i$ and $\tilde{\mathbf{w}}_{i,j}$ denotes the $j$th column corresponding weight matrix $\tilde{\mathbf{W}}_i$. Therefore the optimisation problem over weight matrix $\tilde{\mathbf{W}}_i$ for each LANDO model corresponding to parameter instance $\bm{\mu}_i$ can be written as,
\begin{equation}\label{eq:p_optimization}
    \argmin_{\tilde{\mathbf{W}}_i} \lVert \mathbf{Y}_i - {\tilde{\mathbf{W}}_i} k(\tilde{\mathbf{X}},\mathbf{X}_i) \rVert_F, \quad \text{for $\, i = 1,\cdots,N_{\mu}$}.
\end{equation}
After the computation of the $\tilde{\mathbf{X}}$ and $\tilde{\mathbf{W}}$ matrices, the LANDO surrogate can be finally expressed as,
\begin{equation}\label{eq:pLANDO_indi}
    \mathbf{f}_i(\mathbf{x}) = \tilde{\mathbf{W}}_i  k(\tilde{\mathbf{X}}_i, \mathbf{x}),
\end{equation}
where $\tilde{\mathbf{X}}_i$ is the sparse dictionary of parameter instance $\bm{\mu}_i$ and the corresponding weight matrix is computed via $\tilde{\mathbf{W}}_i = \mathbf{Y}_i k(\tilde{\mathbf{X}}_i, \mathbf{X}_i)^{\dagger}.$ The aforementioned procedure is viewed as the offline phase of parametric LANDO where all the necessary ingredients for prediction in the online phase are prepared.

\subsection{Online stage}
During online stage, the prediction for $\mathbf{x}(t^*,\bm{\mu}^*)$ at desired time $t^*$ given an arbitrary parameter instance $\bm{\mu}^* \in \mathbb{P}$ is performed. The online stage starts from generating a series of data $\{\mathbf{x}(t^*,\bm{\mu}_i)\}_{i=1}^{N_{\mu}}$ based on the constructed LANDO surrogate models for each $\bm{\mu}_i$ at time instant $t^*$. In time-continuous cases, the LANDO surrogate models reflect the rate of change of states over time and $\mathbf{x}(t^*,\bm{\mu}_i)$ can be obtained by time integration up to time $t^*$,
\begin{equation}\label{eq:ds_surrogate}
    \frac{\mathrm{d}\mathbf{x}(t,\bm{\mu}_i)}{\mathrm{d}t} = \mathbf{f}_i(\mathbf{x}), \quad \text{for $\, i = 1,\cdots,N_{\mu}$},
\end{equation}
with a given initial state of the system. 

The further prediction for $\bm{\mu}^*$ can be realised by learning a mapping $\mathcal{M}^{t^*}: \mathbb{P} \rightarrow \mathbb{R}^N$ between parameter $\bm{\mu}$ and the state $\mathbf{x}_{t^*}(\bm{\mu}) = \mathbf{x}(t^*,\bm{\mu}) $ of time instant $t^*$. A variety of techniques, such as polynomial regression \cite{ostertagova2012modelling}, Gaussian process \cite{williams2006gaussian}, or deep neural networks \cite{goodfellow2016deep}, can be applied for such a regression problem. In this work, a deep neural network (DNN) is employed to approximate the mapping based on the data $\{\mathbf{x}(t^*,\bm{\mu}_i)\}_{i=1}^{N_{\mu}}$ considering its capability of approximating the nonlinearity and complexity
of a latent function. 

In general, a DNN is composed of an input layer, an output layer and multiple hidden layers. Each hidden layer consists of several neurons followed by an activation function responsible for the nonlinearity. The output of a hidden layer can be expressed as,
\begin{equation}\label{eq:nn_layer}
 \mathbf{z}^{(i+1)} = \sigma\left(\bm{\omega}^{(i)} \mathbf{z}^{(i)}+\mathbf{b}^{(i)}\right),
\end{equation}
where $\bm{\omega}^{i}$, $\mathbf{b}^{i}$ and $\sigma$ denote the weights, bias and activation function of the layer respectively, while $\mathbf{z}^{(i)}$ and  $\mathbf{z}^{(i+1)}$ are the input and output of this layer. For notation simplicity, we denote $i$th hidden layer as $\varphi_i(\cdot)$. Subsequently, a DNN with an architecture of $L$ number of hidden layers to learn the mapping $\mathcal{M}^{t^*}$ can be expressed as a composition function,
\begin{equation}\label{eq:nn}
 \tilde{\mathcal{M}}_{\bm{\theta}}(\bm{\mu}) = \varphi_L \circ \varphi_{L-1} \cdots \circ \varphi_1(\bm{\mu}),
\end{equation}
where $\bm{\theta} = (\bm{\omega}_1,\cdots,\bm{\omega}_L,\mathbf{b}_1,\cdots,\mathbf{b}_L)$ collects all the trainable parameters in the DNN. Note that the last layer (output layer) $\varphi_L$ is typically a linear layer without activation. DNN can be subsequently trained via minimising the empirical loss between its prediction and the ground truth data,
\begin{equation}
\bm{\theta}^* =    \argmin_{ \bm{\theta}} \mathbb{E}_{\bm{\mu}}\big[ \| \mathbf{x}_{t^*}(\bm{\mu})-\tilde{\mathcal{M}}^{t^*}_{\bm{\theta}}(\bm{\mu})\|^2_2\big] \approx 
    \argmin_{ \bm{\theta}} \frac{1}{N_{\mu}} \sum_{i=1}^{N_{\mu}} \| \mathbf{x}_{t^*}(\bm{\mu}_i)-\tilde{\mathcal{M}}^{t^*}_{\bm{\theta}}(\bm{\mu}_i)\|^2_2,
\end{equation}
where the superscript $t^*$ on $\tilde{\mathcal{M}}^{t^*}_{\theta}(\bm{\mu}_i)$ denotes that the DNN is trained by the data $\{\mathbf{x}_{t^*}(\bm{\mu})\}_{i=1}^{N_{\mu}}$ of time instant $t^*$. For the prediction at a different time instant, a distinct DNN needs to be trained based on the particular data of that time instant. The desired mapping $(t,\bm{\mu}) \mapsto \bm{x}(t,\bm{\mu})$ as stated in Section~\ref{sec:ps} is now reformulated and characterised by the DNN $\tilde{\mathcal{M}}_{\bm{\theta}}^{t}(\bm{\mu})$. Therefore, the final approximation of $\mathbf{x}(t^*,\bm{\mu}^*)$ can be achieved by $\tilde{\mathcal{M}}_{\bm{\theta}^*}^{t^*}(\bm{\mu}^*)$.

Note that for high-dimensional dynamical systems, e.g. the one stemming from spatial-discretised PDEs, training a DNN will be time-consuming and inefficient considering that the dimension of the state $N$ could possibly reach $10^3$ or even more. Therefore, we propose that for such systems, dimensionality reduction techniques, e.g. proper orthogonal decomposition, can be employed to effectively reduce the dimensionality of the state before learning the mapping with a DNN. The generated data $\mathbf{x}_{t^*}(\bm{\mu}_i)$ can be approximated as follow,
\begin{equation}
    \mathbf{x}_{t^*}(\bm{\mu}) \approx \Phi \mathbf{x}^r_{t^*}(\bm{\mu}),
\end{equation}
where $\mathbf{x}^r_{t^*}(\bm{\mu}) \in \mathbb{R}^n$ denotes the reduced representation of $\mathbf{x}_{t^*}(\bm{\mu}_i)$ and $\Phi = [\phi_1|\cdots|\phi_n] \in \mathbb{R}^{N\times n}$ is the matrix consisting of reduced bases $\phi_i$ as columns. The reduced bases are constructed from columns of the left orthonormal matrix via singular value decomposition (SVD) on the snapshot matrix $[\mathbf{x}_{t^*}(\bm{\mu}_1)|\cdots|\mathbf{x}_{t^*}(\bm{\mu}_{N_{\mu}})]$ by Eckart-Young theorem \cite{eckart1936approximation}. The mapping $\mathcal{M}^{t^*}: \mathbb{P} \rightarrow \mathbb{R}^n$ is subsequently no longer between the parameter $\bm{\mu}$ and the state $\mathbf{x}_{t^*}(\bm{\mu})$ but between the parameter $\bm{\mu}$ and the reduced state $\mathbf{x}^r_{t^*}(\bm{\mu})$, and $n \ll N$. The prediction of $\tilde{\mathcal{M}}_{\bm{\theta}^*}^{t^*}(\bm{\mu}^*)$ for the reduced state will be recovered to the full state with the operation $\Phi \tilde{\mathcal{M}}_{\bm{\theta}^*}^{t^*}(\bm{\mu}^*)$. The offline and online stage of parametric LANDO can be summarized in Figure \ref{fig:diagram}.

\section{Numerical examples}\label{sec:example}

In this section, the performance of the parametric LANDO framework is showcased through three numerical examples, the Lotka-Volterra model, the heat equation and the Allen-Cahn reaction-diffusion equation. The parameter spaces are defined respectively for each example and the samples are collected using Latin hypercube sampling. The samples are subsequently separated into a training dataset $\mathcal{D}_{\mathrm{train}}$, a validation dataset $\mathcal{D}_{\mathrm{valid}}$ and a test dataset $\mathcal{D}_{\mathrm{test}}$. The parametric framework for each numerical example is trained on $\mathcal{D}_{\mathrm{train}}$, while the validation dataset $\mathcal{D}_{\mathrm{valid}}$ is used to prevent overfitting in the DNN training. To evaluate the performance of parametric LANDO, the quality of predictions is measured over the test dataset $\mathcal{D_{\mathrm{test}}}$. The relative $L_2$ error between the prediction $\hat{\mathbf{x}}(t^*,\bm{\mu}^*)  = \tilde{\mathcal{M}}_{\bm{\theta}^*}^{t^*}(\bm{\mu}^*)$ and the reference ground truth $\mathbf{x}(t^*,\bm{\mu}^*)$ at the desired time instant $t=t^*$ with parameter configuration $\boldsymbol{\mu}^* \in \mathcal{D}_{\mathrm{test}}$ is written as,
\begin{equation}\label{relative_error}
\epsilon\left(t^*,\bm{\mu}^*\right) =\frac{\left\|\mathbf{x}(t^*, \bm{\mu}^*)-\hat{\mathbf{x}}(t^*, \bm{\mu}^*)\right\|_{2}}{\left\|\mathbf{x}(t^*, \boldsymbol{\mu}^*)\right\|_{2}}.
\end{equation}
To evaluate the overall predictive performance of the framework over the entire test dataset, mean and standard deviation of the relative $L_2$ error over all samples in $\mathcal{D}_{test}$ are estimated, 
\begin{equation}\label{mean_rel_err}
\bar{\epsilon}_{t^*} = \frac{1}{N_{\mu}^*}\sum_{i=1}^{N_{\mu}^*} \epsilon(t^*,\bm{\mu}^*_i) \quad \text{and}  \quad  s(\epsilon_{t^*}) = \Bigg[\frac{1}{N_{\mu}^*}\sum_{i=1}^{N_{\mu}^*} \big(\bar{\epsilon}_{t^*} - \epsilon(t^*,\bm{\mu}^*_i) \big)^2 \Bigg]^{\frac{1}{2}},
\end{equation}
where $N_{\mu}^*$ is the total number of samples  in the test dataset $\mathcal{D}_{\mathrm{test}} = \{\bm{\mu}^*_i\}_{i=1}^{N_{\mu}^*}$. 

\subsection{Lotka-Volterra model}

The Lotka-Volterra model is commonly used to simulate the population dynamics between two species in an ecosystem \cite{wangersky1978lotka}. This is a demonstative example of low-dimensional dynamical system and it can be expressed as:
\begin{equation}\label{Lotka_Volterra}
    \begin{dcases}
        \ \frac{\mathrm{d}x_1}{\mathrm{d}t} = \alpha x_1 - \beta x_1 x_2 \\
        \ \frac{\mathrm{d}x_2}{\mathrm{d}t} = \delta x_1x_2 - \gamma x_2,
    \end{dcases}
\end{equation}
where $x_1$ and $x_2$ refer to the population of prey and predator respectively. The parameters $\alpha$ and $\gamma$ govern the growth rate of predator and prey, while the parameters $\beta$ and $\delta$ feature the influence of the opponent species.  
The parameters $\gamma \text{ and } \delta$ are fixed as $\gamma=0.2 \text{ and } \delta=0.0025$.
The state of the system is denoted as a collection of $\mathbf{x} = [x_1, x_2]^{\top} \in \mathbb{R}^2$. We present two scenarios, one with varying parameter $\bm{\mu} = \alpha$ and another with varying parameters $\bm{\mu} = (\alpha, \beta)$.
% where $\alpha \text{ and } \beta$ are varied within the intervals $[0.015, 0.1] \text{ and } [0.0012, 0.0022]$ respectively.
For both scenarios, the initial condition of the system is $[80, 20]^{\top}$. Note that the dynamics $\mathbf{F}$ is independent of the initial condition, meaning that the surrogate model of dynamics can be applied to predict the state of the system starting from different initial conditions. In the experiments, the snapshots used for training are generated from solving the system using an implicit backward differentiation formula with 600 equidistant timesteps over  $\mathcal{T}_{train} = [0, 400]$.

In the first scenario, $\alpha$ is considered to vary from 0.015 to 0.1 and $\beta$ is fixed as 0.002. To estimate the performance of the framework, 500 test parameter instances are sampled from the parameter space $[0.015, 0.1]$. The threshold $\nu$ of the ALD test as shown in Algorithm~\ref{alg:lando} is set as $10^{-6}$, resulting in an average dictionary size of 6. A quadratic kernel is used in the kernel representation of LANDO considering the quadratic dynamics of the Lotka-Volterra model. A DNN with 3 hidden layers, each containing 32 neurons, is employed to learn the mapping $\mathcal{M}^{t^*}$. The snake activation function \cite{ziyin2020neural} is used to capture the periodic behaviour of parametric problems, demonstrating improved reliability and generalization compared to other commonly used activation functions, such as ReLU.

\begin{figure}[tb!]
    \centering
    \includegraphics[width=\linewidth]{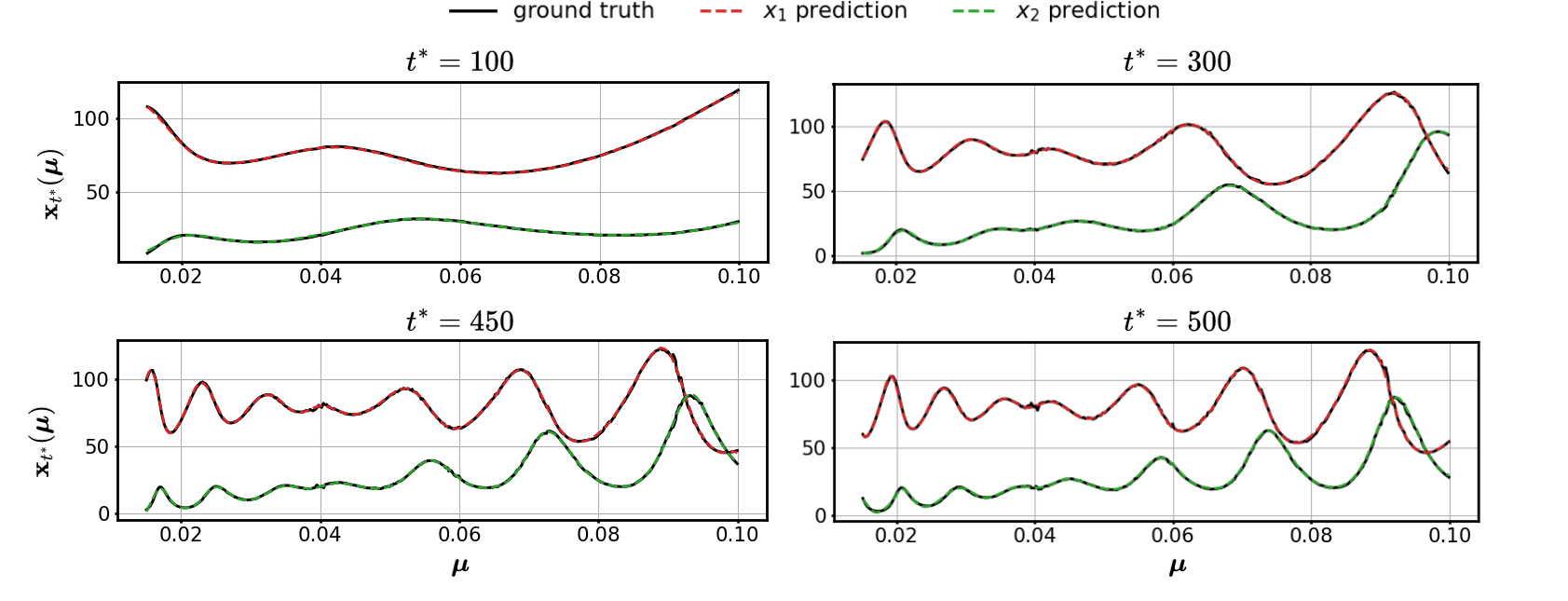}
    \caption{A comparison between the parametric LANDO predictions and the ground truth
    of the Lotka-Volterra model with varying parameter $\alpha$ at time instants 100, 300, 450 and 500. The dynamical system is initiated from $\mathbf{x}_0=[80, 20]^{\top}$.}
    \label{fig:lotka-volterra_1d}
\end{figure}

\begin{figure}[tb!]
    \centering
    \includegraphics[width=\linewidth]{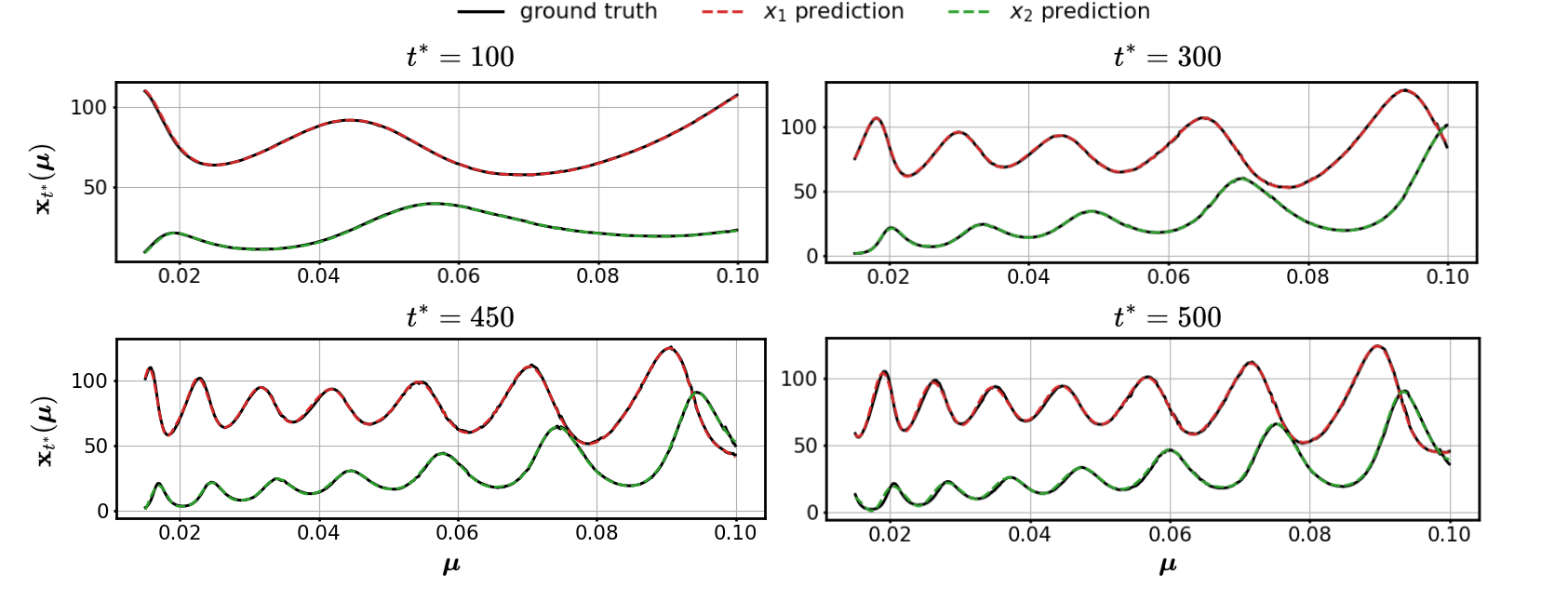}
    \caption{A comparison between the parametric LANDO predictions and the ground truth
    of the Lotka-Volterra model varying parameter $\alpha$ at time instants 100, 300, 450 and 500. The dynamical system is initiated from $\mathbf{x}_0=[70, 20]^{\top}$.}
    \label{fig:lotka-volterra_1d_diffinit}
\end{figure}

The predictions of the parametric surrogate model at four different time instants are demonstrated in Figure~\ref{fig:lotka-volterra_1d}. It can be observed that parametric LANDO achieves a high level of accuracy, with predictions aligning with the ground truth across the parameter space. This alignment persists even for time instants $t^*> 400$, which lie outside $\mathcal{T}_{train}$. The solution manifolds are smooth and show more oscillations when $t^*>100$. The DNN effectively approximates the dynamics, even in cases when the magnitude of oscillations varied across the parameter space. Similarly, Figure~\ref{fig:lotka-volterra_1d_diffinit} demonstrates the predictions of parametric LANDO initiated from a different initial condition. The results further confirm the efficacy of the parametric surrogate model, demonstrating its capability to generalize to varying initial conditions. The snake activation function employed in the DNN captures the periodic behaviour well.

\begin{figure}[tb!]
    \centering
    \includegraphics[width=0.88\linewidth]{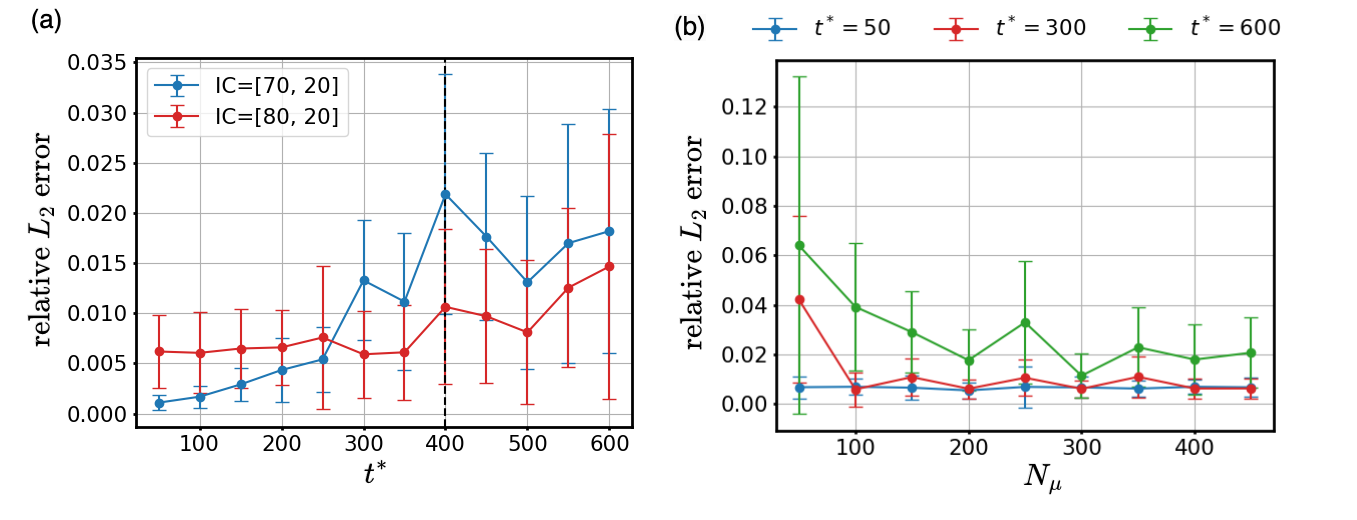}
    \caption{(a) Mean and standard deviation of the relative $L_2$ error of parametric LANDO prediction for time instants from 50 to 600. (b) Mean and standard deviation of the relative $L_2$ error of parametric LANDO prediction with different size of the training dataset at time instants 50, 300 and 600.}
    \label{fig:lotka-volterra_1d_errors}
\end{figure}

To further evaluate the performance of the parametric surrogate model, the mean and the standard deviation of the relative $L_2$ error over time of the two experiments are shown in Figure \ref{fig:lotka-volterra_1d_errors}(a). The vertical dashed line separates the time window covered by $\mathcal{T}_{train}$ from the extrapolation period beyond the training data. The prediction error increases over time, primarily due to the accumulation and propagation of discrepancies within the LANDO framework during the time integration process. Moreover, as shown in Figures \ref{fig:lotka-volterra_1d} and \ref{fig:lotka-volterra_1d_diffinit}, with the evolution of time, the solution manifold becomes more complex and it is more challenging for the DNN to learn the mapping $\mathcal{M}^{t^*}$. However, 
the mean of the relative $L_2$ error is lower than $2\%$ for the majority of time instants and remains below $2.2\%$ even when the dynamics evolved 1.5 times beyond the time window of the training. It can be also observed that the standard deviation of the relative $L_2$ error increases over time. Figure \ref{fig:lotka-volterra_1d_errors}(b) presents the mean and the standard deviation of the relative $L_2$ error of parametric LANDO trained by different sizes of $\mathcal{D}_{train}$ at time instants 50, 300, and 600. As expected, the highest errors are observed when the training set consists of only 50 parametric instances. With the increase of $N_{\mu}$, both the mean and the standard deviation of the relative $L_2$ error decrease. Due to the error accumulation and propagation over time, the mean and the standard deviation of the relative $L_2$ error for time instant 600 is higher than the other two.

\begin{figure}[tb!]
    \centering
    \includegraphics[width=0.87\linewidth]{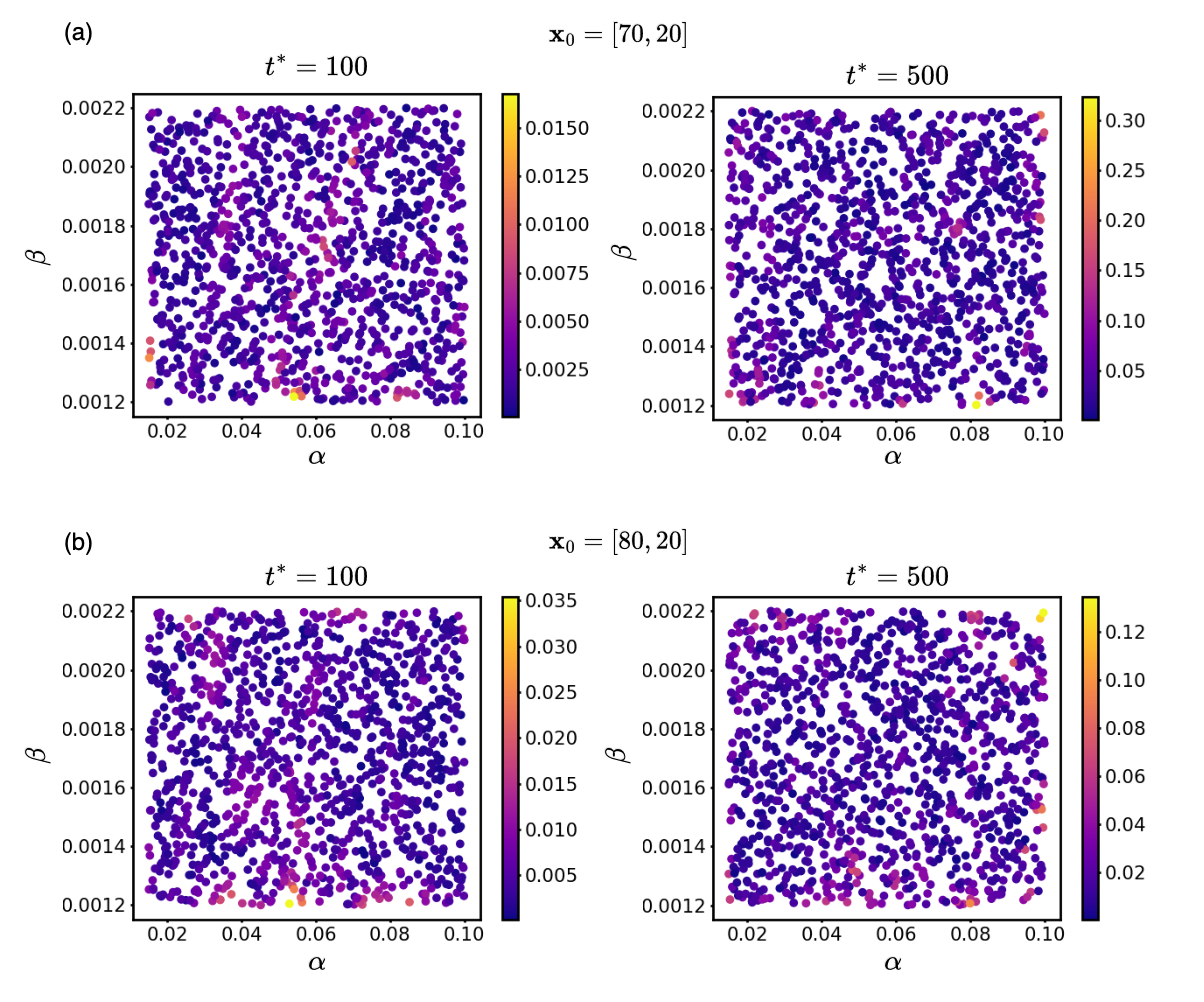}
    \caption{The relative $L_2$ error of parametric LANDO prediction for the Lotka-Volterra model at time instants 100 and 500. Both $\alpha$ and $\beta$ varied simultaneously. (a) with initial condition $x_0=[70,20]^{\top}$, (b) with initial condition $x_0=[80,20]^{\top}$}.
    \label{fig:lotka-volterra_2d}
\end{figure}

\begin{figure}[tb!]
    \centering
    \includegraphics[width=0.88\linewidth]{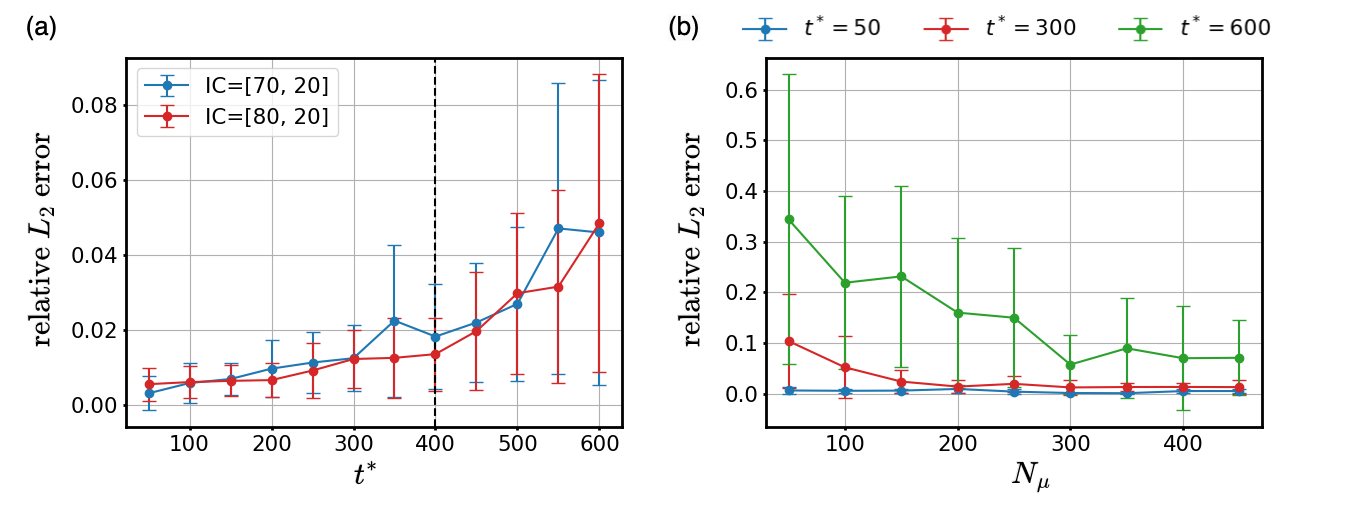}
    \caption{(a) Mean and standard deviation of the relative $L_2$ error of parametric LANDO prediction for time instants from 50 to 600. (b) Mean and standard deviation of the relative $L_2$ error of parametric LANDO prediction with different size of the training dataset at time instants 50, 300 and 600.}
    \label{fig:lotka-volterra_2d_errors}
\end{figure}

In the second scenario, both $\alpha$ and $
\beta$ are considered to vary within the intervals $[0.015, 0.1] \text{ and } [0.0012, 0.0022]$ respectively. The training dataset $\mathcal{D}_{train}$ consists of 560 parametric instances, while the test dataset $\mathcal{D}_{test}$ consists of 1200 test samples, both of which were generated using Latin hypercube sampling. The relative $L_2$ error associated with each parametric instance of $\mathcal{D}_{test}$ at two different time instants and for two different initial conditions is visualised in Figure~\ref{fig:lotka-volterra_2d}. For both initial conditions, the parametric LANDO predicts the parametric dynamics well at $t^*=100$, with errors ranging from 1\% to 3\%. However, the error increase significantly as the prediction approached $t^*=500$. Most of the instances exhibiting high relative error are located close to the boundaries of the parameter space. This aligns with expectations, since those regions are typically not well-covered by the training data, thereby affecting the performance of the DNN for prediction. 

The mean and standard deviation of the parametric LANDO predictions are shown in Figure~\ref{fig:lotka-volterra_2d_errors}(a). Both mean and standard deviation increase over time, consistent with observations from the previous scenario. The mean relative error is approximately 0.5\% at time instant 50 and reached around 5\% when time evolved to 600. Figure~\ref{fig:lotka-volterra_2d_errors}(b) illustrates the relative errors of parametric LANDO prediction with different numbers of training data at time instants 50, 300 and 600. The error at time instant 600 is significantly reduced with a larger training dataset, while the predictive performance of parametric LANDO at time instant 50 is less sensitive to changes in dataset size due to its relatively simpler dynamics. Furthermore, attributed to the increased complexity of the mapping $\mathcal{M}^{t^*}_{\theta^*}$ resulting from a higher input dimension, the average error in this scenario is higher compared to the previous one.

\subsection{Heat equation}

\begin{figure}[tb!]
    \centering
    \includegraphics[width=0.92\linewidth]{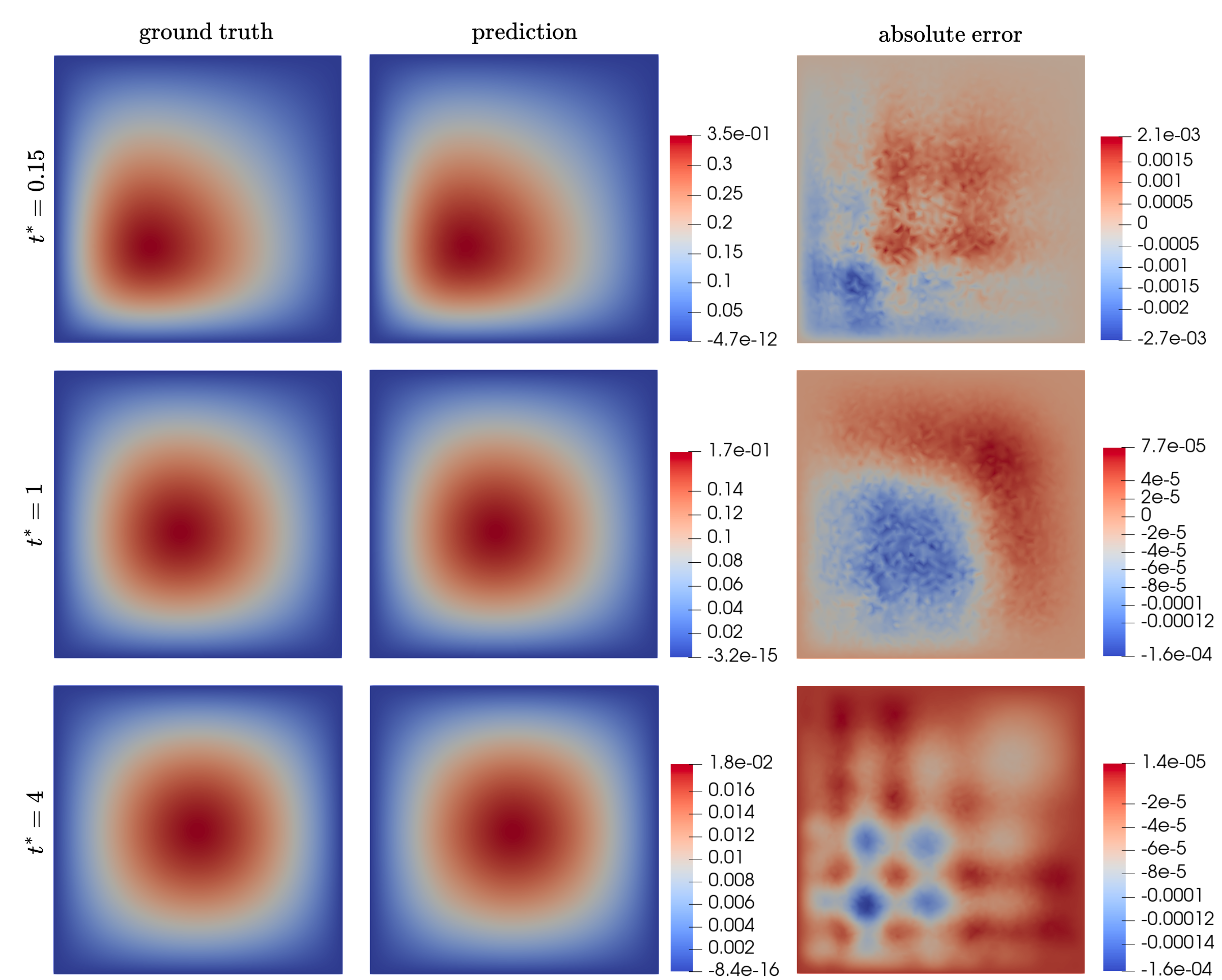}
    \caption{A comparison between the finite element simulation results and the predictions of parametric LANDO at time instants 0.15, 1 and 4 for heat equation. The diffusion coefficient associated to the figures is $D = 0.741$}.
    \label{fig:heat_eq_plando}
\end{figure}

\begin{figure}[tb!]
    \centering
    \includegraphics[width=0.95\linewidth]{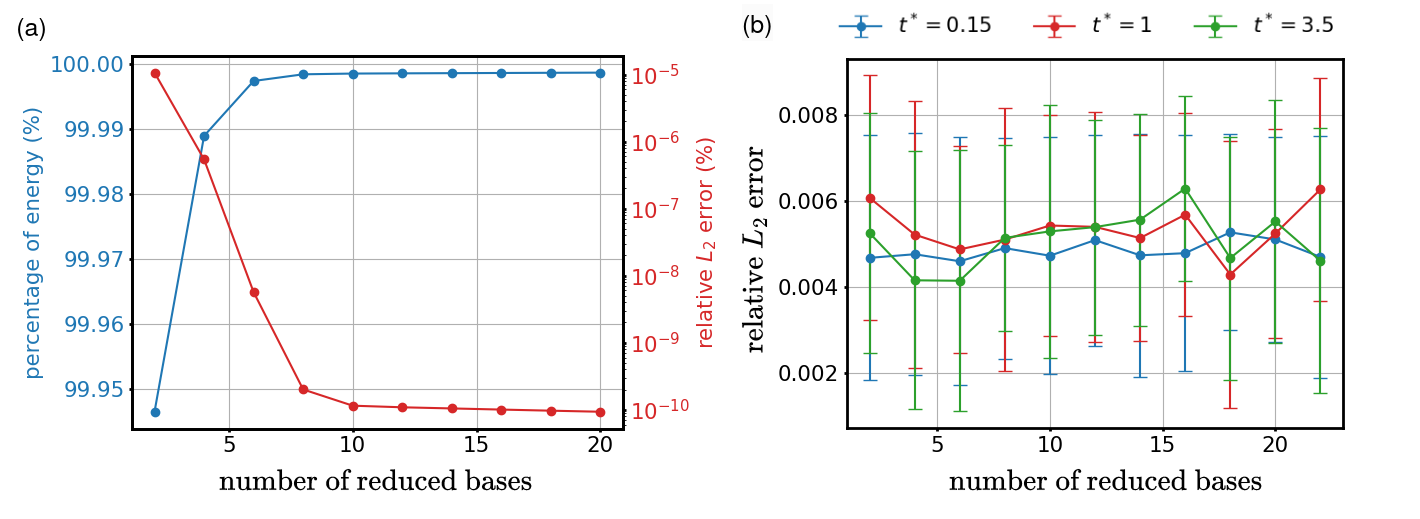}
    \caption{(a) Percentage of energy captured by the number of the reduced bases and the corresponding projection error of POD. (b) Mean and standard deviation of parametric LANDO when POD is performed with several different truncation thresholds. Results from three time instances. }
    \label{fig:heat_eq_pod_experiments}
\end{figure}

\begin{figure}[tb!]
    \centering
    \includegraphics[width=0.88\linewidth]{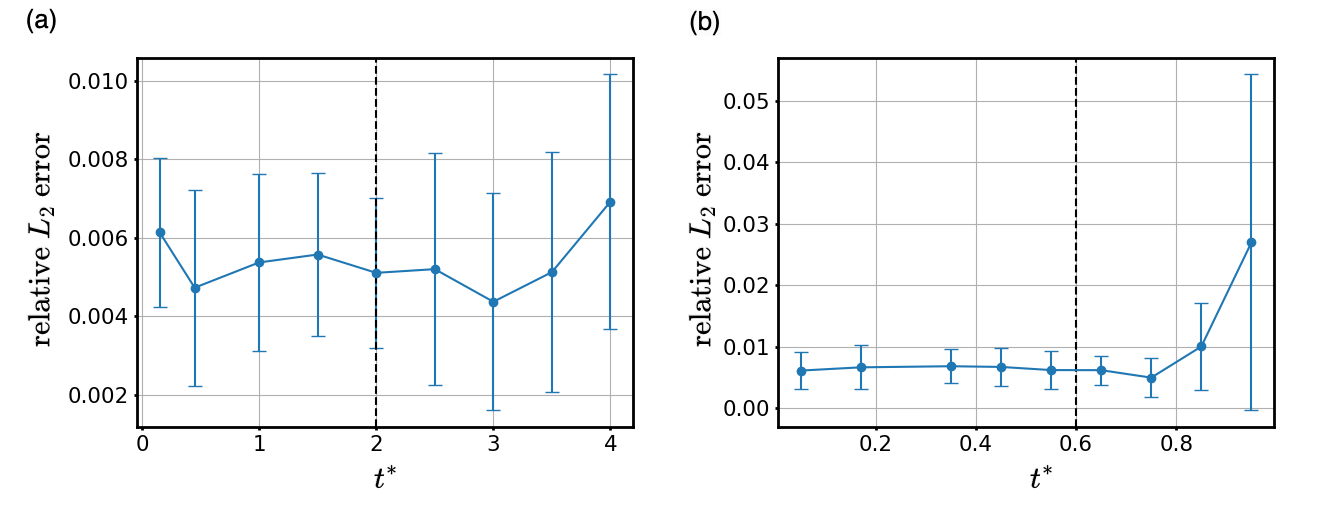}
    \caption{(a) Mean and standard deviation of the relative $L_2$ error of parametric LANDO for time instants from 0.15 to 4, over all parametric instances $\boldsymbol{\mu}_i \in \mathcal{D}_{test}$. The dashed line refers to the training window of LANDO. Parametric LANDO is applied to the heat equation. (b) Mean and standard deviation of the relative $L_2$ error of parametric LANDO for time instants from 0.05 to 0.95, over all parametric instances $\boldsymbol{\mu}_i \in \mathcal{D}_{test}$. The dashed line refers to the training window of LANDO. Parametric LANDO is applied to the Allen-Cahn equation.}
    \label{fig:heat_eq_allen_cahn_errors}
\end{figure}

In this example, we present a two-dimensional parametric heat equation along with its initial and boundary conditions,
\begin{equation}\label{Heat_2d_eq}
\begin{dcases}
\ \frac{\partial u}{\partial t} = D \Big(\frac{\partial^2 u}{\partial x^2}+ \frac{\partial^2 u}{\partial y^2} \Big), \quad(x, y, t) \in(0, 5) \times (0, 5) \times (0, 4] \\
\ u(0, y, t)=u(5, y, t) = 0  \\
\ u(x, 0, t)=u(x, 5, t)=0 \\
\ u(x, y, 0) = \tanh\Bigg(\frac{\alpha \sin{(0.2\pi x)}}{1-\alpha \cos{(0.2\pi x)}}\Bigg)\tanh\Bigg( \frac{\alpha \sin{(0.2\pi y)}}{1-\alpha \cos{(0.2\pi y)}}\Bigg), 
\end{dcases}
\end{equation}
where $u(x,y,t)$ denotes the temperature at point $(x,y)$ of time instant $t$. The coefficient $\alpha$ in the initial condition is set to 0.6 and the diffusion coefficient $D$ is considered as the parameter that varies in $[0.5,1]$. Such a PDE problem can be reformulated into a dynamical system by discretisation of the spatial domain. In this example, a triangular mesh is generated using the open source finite element solver \textit{FreeFEM++} \cite{MR3043640}. The state of the derived dynamics system consists of the temperature at each vertex of the mesh. Due to the large number of vertices, this system is characterized as high-dimensional, necessitating the integration of POD into the parametric framework for dimensionality reduction. To construct the training, validation and test dataset, 150, 50 and 100 parameter instances are sampled from $[0.5, 1]$ respectively and fed to \textit{FreeFEM++} to perform simulation for data generation. LANDO is trained by the equidistant snapshots with a time step $\Delta t = 0.01$ over $[0, 2]$. During testing, the parametric surrogate model performs prediction up to $t^* = 4$. Note that in this example, we implement the LANDO algorithm with the time-discrete form as shown in \eqref{eq:ds_discrete}. A linear kernel function is chosen for LANDO with the sparsity threshold set at $10^{-5}$. A DNN with 4 hidden layers, each containing 110 neurons, is employed to learn the mapping between parameters and reduced states. 

A comparison of the prediction of parametric LANDO and ground truth generated from FEM simulations at time instants 0.15, 1 and 4 is shown in Figure~\ref{fig:heat_eq_plando}. For all three time instances, parametric LANDO manages to capture the dynamics of the system, with predictions closely aligning with the ground truth. The maximum absolute error between the prediction and ground truth is around $2.1\time 10^{-3}$ at time instant 0.15. The error decreases at subsequent time instances due to the diffusion behaviour of the system, leading to a gradual reduction in temperature over time. Figure~\ref{fig:heat_eq_pod_experiments} investigates the performance of parametric LANDO with respect to the choice of POD. The mean and standard deviation of the parametric LANDO prediction with different numbers of reduced bases are demonstrated. Notably, the snapshots are well approximated even with only four bases, covering approximately $99.99\%$ of the energy while maintaining a negligible approximation error of around $10^{-6}$. This considerable reduction in dimension significantly enhances the computational efficiency of the parametric surrogate model, as the output of the DNN corresponds to the reduced state with a dimension of four rather than the full state. A further improvement in POD can hardly contribute to the prediction performance, as shown in Figure~\ref{fig:heat_eq_pod_experiments}(b). Moreover, the overall performance of parametric LANDO is demonstrated in Figure~\ref{fig:heat_eq_allen_cahn_errors}(a). It can be observed that the relative $L_2$ error remains below $1\%$, even for the prediction at time instant 4.

\subsection{Reaction-diffusion equation}

\begin{figure}[tb!]
    \centering
    \includegraphics[width=\linewidth]{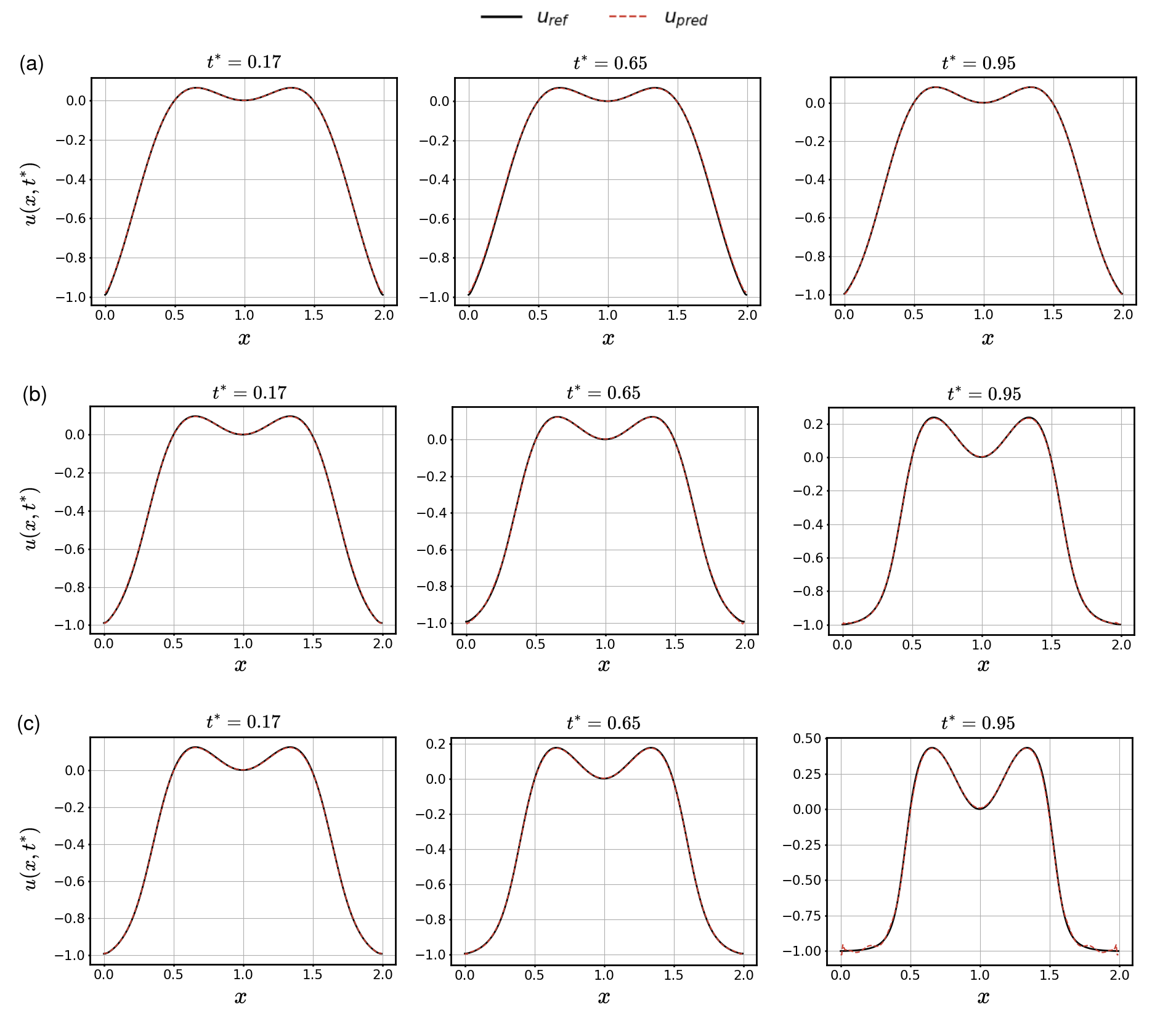}
    \caption{A comparison between the prediction of parametric LANDO and the reference solution of the Allen-Cahn equation at three different parametric and time instances. (a) $\bm{\mu} = (0.00074, 1.935)$. (b) $\bm{\mu} = (0.000906,3.009)$. (c) $\bm{\mu} = (0.000579,3.807)$.}
    \label{fig:allen_cahn_lando}
\end{figure}

\begin{figure}[tb!]
    \centering
    \includegraphics[width=0.95\linewidth]{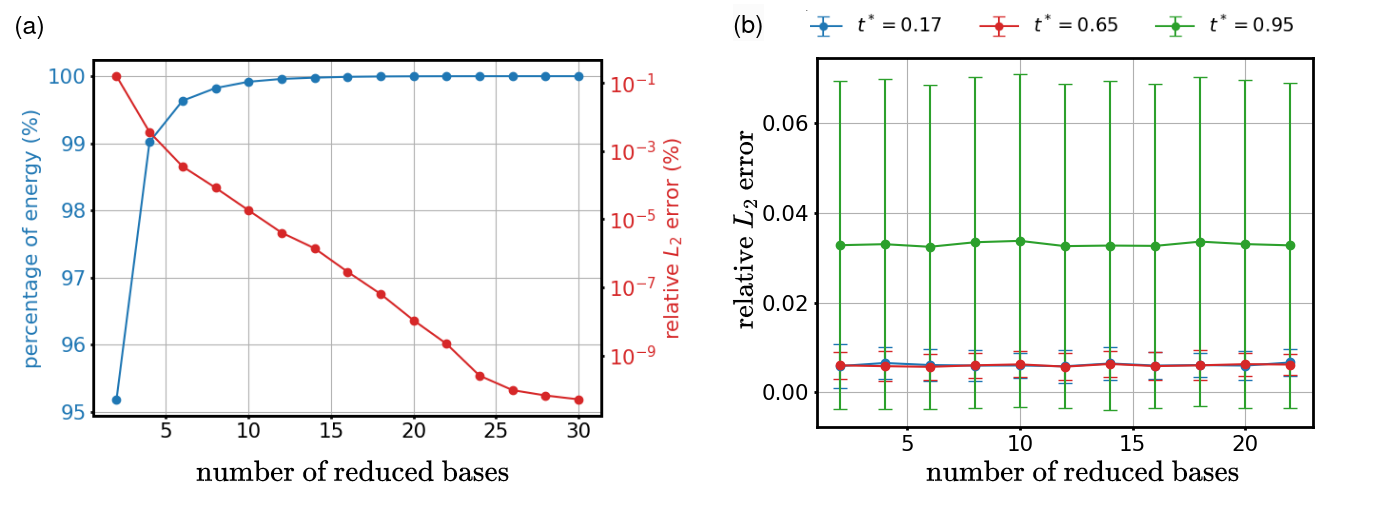}
    \caption{(a) Percentage of energy captured by the number of the reduced bases and the corresponding projection error of POD. (b) Mean and
standard deviation of parametric LANDO when POD is performed with several different truncation thresholds. Results from three time instances.}
    \label{fig:allen_cahn_podexperiments}
\end{figure}

Reaction-diffusion models were originally developed for simulating chemical reactions and have been adapted for a wide range of disciplines such as biology, physics and ecology \cite{allen1975coherent,tersian2001periodic,kondo2010reaction}. The Allen-Cahn equation is one of the reaction-diffusion equations particularly useful for modelling phase separation processes \cite{moelans2008introduction, shen2010numerical}. It is expressed as follows,
\begin{equation}\label{Allen_Cahn_eq}
\begin{dcases}
 \  \frac{\partial u}{\partial t}-\lambda \frac{\partial^2 u}{\partial x^2}+\varepsilon f(u)=0, \quad(x, t) \in(-1,1) \times(0,1] \\
\ u(-1, t)=u(1, t) =-1 \\
\ u(x, 0)=x^2 \cos (\pi x) \\
\ f(u) = u^3-u,
\end{dcases}
\end{equation}
where $u(x,t)$ denotes the state variable. $\lambda$ refers to the diffusion coefficient, and $\varepsilon$ scales the nonlinear reaction term $f(u)$. The dynamic of the parametric system is characterised by two parameters, i.e. $\bm{\mu} = (\lambda,\varepsilon)$ varying within $[0.0001, 0.001] \times [0.5, 4]$. The training, validation and test datasets consist of 400, 150 and 550 samples, respectively, with their corresponding snapshots generated using the finite difference method. The one-dimensional spatial domain is discretised into 249 equidistant intervals. POD is again applied to reduce the dimension of the state before training the DNN. LANDO models are utilized to approximate the dynamics with a time-discrete setting and linear kernel. The same DNN configuration of the heat equation is adopted. The sparsity threshold is set to be $\nu = 10^{-6}$. 

The ground truth and the parametric LANDO predictions for three different parameter instances at time instants 0.17, 0.65 and 0.95 are illustrated in Figure~\ref{fig:allen_cahn_lando}. The results indicate that parametric predictions of the proposed method effectively capture the dynamics of the system, even at time beyond the training time window $[0,0.6]$. Minor errors are observed close to the boundary of the domain for the prediction corresponding to the parameter instance $\bm{\mu} = (0.000579,3.807)$ at time instance 0.95. The overall performance of the parametric surrogate model is reported in Figure~\ref{fig:heat_eq_allen_cahn_errors}(b). A significant increase in the average relative error and standard deviation can be observed when the dynamics evolved beyond $t^*=0.8$. Additionally, the influence of POD is also evaluated. Figure~\ref{fig:allen_cahn_podexperiments}(a) demonstrates the cumulative energy covered by reduced bases and the associated discrepancy, indicating that ten reduced bases are required to cover $99.99\%$ cumulative energy. Similarly to the findings with the heat equation, the performance of the parametric prediction barely improves with an increase in the number of reduced bases, as shown in Figure~\ref{fig:allen_cahn_podexperiments}(b).

% After further investigation it was found that the deteriorated performance occurs due to the inability of non-parametric LANDO to emulate the system dynamics for several parametric configurations $\left(\lambda_i, \varepsilon_i\right)$. This is also a possible explanation for the increased standard deviation in $t^*=0.85$ and $t^*=0.95$, since LANDO can approximate sufficiently well the system dynamics only for a subset of the parameter space. As a result of that, the prediction of parametric LANDO cannot be enhanced for these time instances, since the DNN already learns the mapping on erroneous data. 

\section{Discussion and conclusion}\label{sec:discussion}
In this work, a parametric extension of the LANDO algorithm is proposed. To enable parametric prediction, a series of 
LANDO models are constructed for each parameter instance from the training dataset and used for generating new data at a desired time instant. A deep neural network is subsequently applied to learn the mapping between the parameters and the states for that time instant. For high-dimensional dynamical systems, dimensionality reduction techniques can be applied. The effectiveness of the proposed framework is presented with three numerical examples, and all three cases have demonstrated the decent predictive performance of the framework. The proposed framework can be applied in many-query scenarios, such as uncertainty quantification or design optimisation, where repeated evaluations of a dynamical system are required.  

The error of the proposed framework mainly originates from three perspectives: LANDO prediction, POD and DNN approximation. The LANDO models are used to emulate the dynamics of systems with the parameter instances from the training dataset and subsequently generate the training data to learn the mapping $\mathcal{M}^{t^*}$. The performance may be further improved by setting a more rigorous threshold for the ALD test in sparse dictionary construction. Avoiding extrapolation beyond the training time window can also contribute to the prediction performance as the minimisation of \eqref{eq:optimziation_lando} guarantees the minimal residual between training data and LANDO prediction. Note that the minimisation problem \eqref{eq:optimziation_lando} can be extended to multi-steps, also known as roll-out scheme \cite{uy2023operator}. The rollouts enforce the more accurate prediction of LANDO by minimising the residual between predictions over multiple time steps and can make the surrogate models more robust against noise and scarce data. However, such optimisation also will compromise the computational efficiency since the roll-out problems typically take much longer to solve.

To mitigate the curse of dimensionality, the model reduction technique is applied to approximate the solution manifold. Such techniques also inevitably introduced discrepancies. In the second and third examples, POD effectively reduces the dimension of the full state to a reduced state with only a small number of dimensions. For the problems suffering from slowly decaying of Kolmogorov N-width due to the domination of the advection or non-affined parametrisation, nonlinear model order reduction methods, such as kernel-PCA \cite{scholkopf1997kernel} and autoencoder/decoder \cite{romor2023non}, can be applied to mitigate the issue. The performance of the parametric framework also substantially depends on how well the DNN has approximated the latent mapping $\mathcal{M}^{t^*}$. It is important to note that the solution manifold might not be smooth which leads to further challenges in solution map approximation. \textit{A prior} knowledge, such as physics information or geometry features, can be applied to enhance the efficiency in training and performance of the model. For example, snake activation function is adopted in DNN for the Lotka-Volterra example, considering its periodic behaviour. The accuracy of the LANDO model and the application of model order reduction also play important roles since both of them introduce discrepancy and lead to the noise of the data. 

Considering all those discrepancies mentioned above, a quantification of uncertainty would help us to understand the error propagation in model construction and prediction, and further guarantee the credibility of the parametric surrogate model. Either ensemble methods \cite{fasel2022ensemble} or Bayesian schemes \cite{hirsh2022sparsifying, takeishi2017bayesian} can be integrated into the framework and subsequently provide an estimate of distribution/interval rather than a point estimation. We leave this part of the research to our future work.

\section*{Data availability}
All the data and source codes to reproduce the results in this study are available on GitHub at \url{https://github.com/kevopoulosk/Parametric-kernel-based-DMD-using-deep-learning}

% \section{Declaration of Competing Interest}
% The authors declare that they have no known competing financial interests or personal relationships that could have appeared to influence the work reported in this paper

\section*{Acknowledgement}
D.Ye acknowledge the financial support from Sectorplan Bèta (the Netherlands) under the focus area \emph{Mathematics of Computational Science}. We would like to thank Prof. Daan Crommelin and Dr. Giulia Pederzani for the fruitful discussions during the project.

\bibliographystyle{vancouver}
\bibliography{Reference.bib}

\end{document}